# Sprachsynthese - State-of-the-Art in englischer und deutscher Sprache

René Peinl
Hochschule Hof, Alfons-Goppel-Platz 1, 95028 Hof
rene.peinl@hof-university.de

## Abstract

Das Vorlesen von Texten ist eine wichtige Funktion für moderne Computeranwendungen. Es erleichtert nicht nur sehbehinderten Menschen den Zugang zu Informationen, sondern ist auch für nicht eingeschränkte Benutzer ein angenehmer Komfort. In diesem Artikel werden Stand von Forschung und Technik der Sprachsynthese getrennt nach Mel-Spektrogramm-Generierung und Vocoder dargestellt. Den Abschluss bildet eine Übersicht verfügbarer Datensätze für Englisch und Deutsch mit einer Diskussion der Übertragbarkeit der Ergebnisse von EN nach DE.

## Einleitung

Die Qualität der Sprachsynthese (TTS, Text to Speech) hat sich in den letzten Jahren durch den Einsatz tiefer neuronaler Netze (deep learning) erheblich verbessert. Der blecherne Klang früherer TTS Systeme ist einem recht natürlichen Klang gewichen, wie internationale Publikationen insbesondere für die englische Sprache zeigen [1]. Viele der dort verwendeten Modelle sind auch als Open Source Code im Internet verfügbar. Stand der Technik bei der Sprachsynthese ist derzeit der Einsatz eines zweistufigen Modells. Die erste Stufe generiert ein Zwischenprodukt, ein sogenanntes Mel-Spektrogramm (siehe Abbildung 1). Aus dieser macht die zweite Stufe (Vocoder) dann die eigentliche Audio-Datei. Das Training erfolgt üblicherweise mit kurzen Sätzen oder Satzteilen (1-15 s) und erfordert eine präzise Zuordnung von Text zu Audio-Material. Je genauer diese Zuordnung erfolgen kann (Alignment), desto besser der Trainingseffekt. Die besten Ergebnisse werden aktuell mit Trainingsdaten eines einzelnen Sprechers erzielt. 20 Stunden gelten dabei als Minimum, bis zu 50 h werden für gute Modelle verwendet. Die Qualität der Aufnahme ist ebenfalls wichtig für die spätere Qualität der Sprachausgabe. Nebengeräusche und schlechte Mikrofontechnik führen zu schlechteren Ergebnissen.

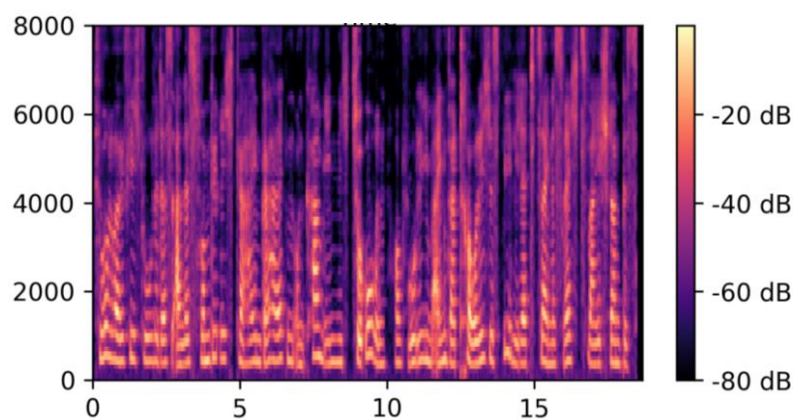

Abbildung 1: Mel-Spektrogramm mit Zeit in s in X-Richtung und Frequenz in Hz in Y-Richtung (Quelle: eigene Darstellung)

Die Qualität der Sprachsynthese lässt sich bisher automatisiert nur schlecht beurteilen. Daher sind in der Literatur überwiegend subjektive Verfahren wie der Mean Opinion Score (MOS) oder (seltener) das MUSHRA Verfahren (Multiple Stimuli with Hidden Reference and Anchor) im Einsatz. Der Vorteil von MUSHRA gegenüber MOS ist, dass für jeden Satz alle zu bewertenden Systeme gleichzeitig in einem Panel dargestellt werden können [2]. Beim MOS werden die Audiomaterialien dagegen nacheinander bewertet [3]. Bei jedem MUSHRA Test werden die Positionen der Systeme im Panel randomisiert. Beim MOS Verfahren muss dagegen die Reihenfolge der Darbietung entsprechend randomisiert werden. Das MOS Verfahren sieht eine Skala von 1 bis 5 vor, wobei eins schlechte und fünf sehr gute Qualität repräsentiert. Bei MUSHRA geht die Skala von 1 bis 100 wobei auch hier die höhere Zahl eine bessere Qualität darstellt. In [4] findet sich ein Vorschlag für zwei neue Qualitätskennzahlen, die automatisiert ermittelbar sind und laut Autoren gut mit den MOS Werten korrelieren. Das könnte eine deutliche Verbesserung der Situation für die Entwicklung von Sprachsynthese bedeuten, da die Forscher schnelle Rückmeldung über die Qualität des Modells erhalten könnten, ohne jedes Mal aufwändige empirische Studien durchführen zu müssen. Mangels unabhängiger Überprüfung der Werte, ist die Brauchbarkeit aber wissenschaftlich noch nicht belegt.

## Vocoder

Der Vocoder generiert meist aus einem Mel-Spektrogramm die Audio-Datei. Da sich das Mel-Spektrogramm basierend auf gegebenen Audio-Dateien problemlos berechnen lässt, kann man Trainingsdaten vergleichsweise einfach erzeugen. Die Qualität aktueller neuronaler Vocoder ist dabei auf dem Niveau menschlicher Sprecher und wird sogar z.T. sogar als überlegen eingestuft [5]. Daher konzentrieren sich die Forschungen in den letzten Jahren auf eine Verbesserung der Performance bei gleichbleibend hoher subjektiver Qualität der Sprachausgabe.

**WaveNet** [6] kann bzgl. der MOS Werte als Referenz gelten. In der Original-Veröffentlichung wurde zwar nur ein MOS von 4,21 berichtet, der im Vergleich zum MOS von 4,55 der Ground Truth zwar gut, aber nicht exzellent ist. In nachfolgenden Veröffentlichungen wurden jedoch Werte nahe der Ground Truth erzielt [7]. Bei der Interpretation dieser Daten muss man berücksichtigen, dass die Modelle nicht immer mit den gleichen Daten trainiert wurden. Daher ist die Vergleichbarkeit nur bedingt gegeben.

**Parallel WaveNet** [8] setzt direkt auf WaveNet auf und gewinnt durch Knowledge Distillation (siehe Tabelle 3) aus einem fertig trainierten WaveNet ein deutlich schnelleres Netz für das Inferencing, welches die Audiodaten in 20-facher Echtzeit generiert (real-time factor RTF 0,05), also in einer Sekunde 20 s Audioausgabe, während Wavenet für 20 s Audioausgabe 1.000 Sekunden brauchte, also fast 17 Minuten (RTF 50). Der Nachteil ist, dass der Trainingsprozess sehr aufwändig ist, weil man zuerst den „Lehrer" trainieren muss und dann den Studenten daraus gewinnt, der im Fall von Parallel WaveNet mit ähnlicher Qualität arbeitet.

**WaveGlow** [9] ist ein Vertreter der neuen Flow-basierten Netzwerke, die im Gegensatz zu den auto-regressiven Netzwerken wie WaveNet gut parallelisierbar sind. Es erreicht dadurch RTF 0,04 auf einer Nvidia Tesla V100 GPU. **FloWaveNet** schlägt in dieselbe Kerbe und kann ähnlich wie WaveGlow in einem einzelnen Trainingsschritt trainiert werden, statt aufwändige Distillation zu erfordern. Die Autoren sprechen von einer WaveNet „vergleichbaren" Qualität, was angesichts von einer Differenz von 0,51 auf der MOS-Skala jedoch beschönigend erscheint. Von der Geschwindigkeit her liegt es auf dem Niveau der Konkurrenz mit einem RTF von 0,05.

**Parallel WaveGAN** [10] kommt ohne Distillation aus und schafft mit nur 1,44 Millionen Parametern eine hohe Qualität von 4,06 MOS und einen RTF von 0,036 auf einer einzelnen V100 GPU. Auffällig ist, dass in der Veröffentlichung der Vergleichs-MOS-Wert von ClariNet leicht über dem Wert in dessen Originalpaper liegt, während der WaveNet Wert mit 3,61 MOS weit unter dem dort berichteten Wert liegt. Mit Parallel WaveGAN kann auch die Trainingszeit deutlich gesenkt werden. [10] berichten von 2,8 Tagen Trainingszeit mit zwei Nividia Tesla V100 GPUs, wohingegen WaveNet 7,4 Tage benötigte und ClariNet sogar 12,7 Tage.

**MelGAN** [11] ist ein Modell, das Generative Adversarial Networks [GAN, 12] für die Sprachsynthese einsetzt. Es stellt schon die zweite Generation von GANs für TTS dar und kommt komplett ohne Distillation Prozess aus. Dabei erreicht es einen MOS Wert von 3,61, was nicht an die Vergleichswerte von 4,05 für WaveNet und 4,11 für WaveGlow heranreicht. Die Weiterentwicklung **Multiband MelGAN** [13] verbesserte diese Werte jedoch schon bald und erreichte mit 4,34 MOS State-of-the-Art Niveau. Mit Tacotron 2 als Mel-Generator erreicht es immer noch sehr gute 4,22 MOS. Gleichzeitig konnte die Anzahl der Parameter auf 1,91 Mio reduziert werden, was einer Reduktion der nötigen Rechenleistung auf ein Sechstel entspricht. Die Ergebnisse müssen jedoch relativiert werden, da sie mit Chinesischer Sprache erzielt wurden, statt in Englisch.

**GAN TTS** [14] arbeitet nicht mit Mel-Spektrogrammen als Ausgangsbasis, sondern mit linguistischen Features und fällt deswegen etwas aus der Reihe. Es erreicht gute MOS Werte durch den Einsatz mehrerer Diskriminatoren als Feedback für den Generator-Teil des GAN, die mit verschiedenen Fenstergrößen arbeiten.

**ClariNet** [15] ähnelt Parallel WaveNet im Knowledge Distillation Ansatz, schafft es aber durch eine Kombination der DeepVoice 3 Architektur [16] für die Generierung von Spektrogrammen mit der distillierten WaveNet

Architektur ein Ende-zu-Ende Netzwerk für die Generierung von Sprachdaten direkt aus Texten zu erzeugen, das trainierbar ist und als Text-to-Wave Architektur bezeichnet wird. Die Autoren räumen aber ein, mit der Kombination nicht ganz die Qualität der einzelnen Bestandteile zu erreichen, weil es im Trainingsprozess zu Beginn schlechte Spektrogramme gibt, die sich auf die Qualität der zweiten Netzhälfte auswirken.

Tabelle 1 Vergleich der Vocoder für Sprachsynthese (GT = Ground Truth, GT-MOS = Differenz der Werte aus den Spalten)

| Modell | Bemerkung | MOS | Vergleich | | GT | GT-MOS |
|---|---|---|---|---|---|---|
| **WaveNet** | 25h Trainingsdaten, 16 kHz | 4,21 | HMM concatenative: 3,86 | | 4,55 | 0,34 |
| **Parallel WaveNet** | 65h Trainingsdaten, 24 kHz | 4,41 | WaveNet: 4,41 | (± 0,00) | 4,55 | 0,14 |
| **WaveGlow** | Trainiert mit LJSpeech | 3,96 | WaveNet: 3,88 | (+0,08) | 4,27 | 0,31 |
| **WaveRNN** | 44 h Trainingsdaten | 4,46 | WaveNet: 4,51 | (-0,05) | | |
| **ClariNet** | Ohne Mel-Spektrogramm | 4,15 | - | | 4,54 | 0,39 |
| **Parallel WaveGAN** | Ergebnisse für Japanisch | 4,06 | ClariNet: 4,21 | (-0,15) | 4,46 | 0,40 |
| **MelGAN** | Trainiert mit LJSpeech | 3,61 | WaveNet: 4,05<br>WaveGlow: 4,11 | (-0,44)<br>(-0,50) | 4,52 | 0,91 |
| **Multiband MelGAN** | Ergebnisse für Chinesisch | 4,22 | MelGAN: 3,87 | (+0,35) | 4,58 | 0,36 |
| **GAN TTS** | Sehr effizient | 4,21 | WaveNet: 4,41 | (-0,20) | 4,55 | 0,24 |
| **FloWaveNet** | Trainiert mit LJSpeech | 3,95 | WaveNet: 4,46 | (-0,51) | 4,67 | 0,72 |
| **Hifi-GAN** | Trainiert mit LJSpeech | 4,36 | WaveNet: 4,02<br>MelGAN: 3,79 | (+0,34)<br>(+0,57) | 4,45 | 0,09 |
| **WaveGrad** | Trainiert mit LJSpeech | 4,55 | MelGAN: 3,95<br>WaveRNN: 4,49 | (+0,56)<br>(+0,06) | 4,55 | 0,00 |

**HiFi GAN** [17] arbeitet wie der Name schon sagt mit Generative Adversarial Networks und kann unterschiedlich parametrisiert werden, um entweder beste Qualität oder auch beste Performance zu erreichen. In der schnellsten getesteten Konfiguration ergibt sich ein Realtime Faktor von 0,08 auf der CPU und 0,001 auf der GPU bei einer gleichzeitigen Reduktion der Qualität um 0,31 MOS. Im Vergleich dazu schafft WaveGlow nur 4,8 auf der CPU und 0,04 auf der GPU (Nvidia V100). In der höchsten Qualitätsstufe sinkt die Performance von HiFi GAN auf immer noch gute 0,7 RTF (CPU) bzw. 0,006 (GPU). In der besten Qualität liegt es nur um 0,09 MOS hinter der Ground Truth (siehe Tabelle 1)

Noch besser schneidet **WaveGrad** ab [18]. Es hat sowohl absolut als auch relativ zur Ground Truth den besten MOS Wert. Es ist auch dahingehend interessant, als es iterativ arbeitet und damit problemlos das Einstellen des Modells zwischen Qualität der Ergebnisse durch mehr Iterationen und Geschwindigkeit der Ergebnisgenerierung mit weniger Iterationen erlaubt. Es schlägt WaveRNN sowohl bei LJSpeech als auch mit einem eigenen Multi-Speaker Datensatz mit 385 Stunden knapp und liegt gleichauf mit den MOS Werten der menschlichen Sprecher. Selbst bei einer Reduktion von 1000 Iterationen auf sechs ist die Qualität der Ergebnisse mit 4,41 MOS noch sehr hoch und erreicht dabei RTF 0,2 auf einer V100 GPU.

Auch Mozilla experimentiert mit TTS, nachdem sie im Bereich Spracherkennung mit dem Common Voice Projekt zum Sammeln von Trainingsdaten in vielen Sprachen und der DeepSpeech Implementierung große öffentliche Aufmerksamkeit erzeugt haben. Mit **LPCNet** [19] schlagen sie ein Modell vor, das auf WaveRNN basiert und bei gleicher Komplexität besser, oder bei gleicher Qualität deutlich effizienter sein soll. Leider wird diese Behauptung im Paper nicht mit MOS-Werten untermauert, sondern nur mit dem weniger gängigen MUSHRA-Verfahren, wo es mit Werten über 80 besser abschneidet als WaveRNN. Nach der Schließung der Abteilung bei Mozilla werden die Arbeiten unter dem Namen **Coqui**[1] von einem Berliner Startup weitergeführt.

Tabelle 1 fasst die publizierten MOS Werte der Vocoder zusammen. Wo nicht anders angegeben wird englische Sprache verwendet. Menge und Qualität der Trainingsdaten wurde dort wo publiziert in der Spalte Bemerkung vermerkt. Da ein direkter Vergleich der MOS Werte unterschiedlicher Studien nicht zulässig ist, sind die Vergleichswerte aus den zitierten Veröffentlichungen mit angegeben.

---

[1] https://github.com/coqui-ai/TTS

# Mel-Spektrogramm Generierung

Die früher üblichen Modelle zur Mel-Spektrogramm-Generierung basierend auf Hidden Markov Modellen (HMM) wurden in den letzten Jahren zunehmend von Deep Learning-basierten Ansätzen verdrängt. Insbesondere **Tacotron** [1] und sein Nachfolger **Tacotron 2** [7] haben zu einem regelrechten Hype der Forschungen im Bereich Sprachsynthese geführt. Während Tacotron noch einen Griffin-Lim Vocoder als zweite Stufe einsetzt und damit nur einen MOS von 3,82 erreicht, gelingt es Tacotron 2 mit einem durchgängigen auf Deep Learning basierten Verfahren einen MOS-Wert von 4,53 zu erzielen, was sehr nahe an dem Wert menschlicher Sprecher liegt (4,58). Dazu verwenden die Forscher eine modifizierte Version von WaveNet als Vocoder.

Während Tacotron auf Recurrent Neural Networks (RNNs) basiert, die häufig für Sprachsynthese eingesetzt werden, gelang es [20] die Transformer Architektur, die aus der Sprachverarbeitung mit Modellen wie BERT [21] bekannt wurde, erfolgreich auf Sprachsynthese anzuwenden und damit ähnliche oder leicht bessere Werte als Tacotron 2 zu erzielen. **Transformer TTS** [20] erzielt einen MOS Wert von 4,39 im Vergleich zu 4,44 der menschlichen Sprecher und liegt damit gleichauf mit Tacotron 2. Da MOS ein subjektiver Wert ist, gibt es leider keine Möglichkeit die Ergebnisse unterschiedlicher Studien direkt zu vergleichen. Daher wird in der Regel Tacotron 2 als Referenz und Benchmark genutzt (siehe Tabelle 2).

Tabelle 2: Vergleich der Sprachsynthese Modelle auf Basis tiefer neuronaler Netze

| Modell | Vocoder | MOS | Vergleich | GT | GT-MOS | Mel GT | Zeit |
|---|---|---|---|---|---|---|---|
| **Tacotron 2** | WaveNet | 4,53 | Tacotron: 4,00 | 4,58 | 0,05 | x | 4,58 s |
| **Transformer TTS** | WaveNet | 4,39 | Tacotron 2: 4,385 | 4,44 | 0,05 | x | 3,26 s |
| **Fastspeech** | WaveGlow | 3,84 | Transformer TTS: 3,88 | 4,41 | 0,57 | 4,00 | 0,16 s |
| **Fastspeech 2** | Parallel WaveGAN | 3,79 | Transformer TTS: 3,79 Tacotron 2: 3,74 | 4,27 | 0,48 | 3,92 | ~0,16 s |
| **AlignTTS** | WaveGlow | 4,05 | Transformer TTS: 4,02 | 4,53 | 0,48 | 4,28 | 0,18 s |
| **Glow-TTS** | WaveGlow | 4,01 | Tacotron 2: 3,88 | 4,54 | 0,53 | 4,19 | ~0,29 s |
| **Flow-TTS** | WaveGlow | 4,19 | Tacotron 2: 4,10 | 4,55 | 0,36 | 4,35 | ~0,20 s |
| **Flowtron** | WaveGlow | 3,66 | Tacotron 2: 3,52 | 4,27 | 0,61 | x | x |
| **TalkNet** | WaveGlow | 3,74 | Tacotron 2: 3,85 | 4,31 | 0,57 | 4,04 | ~0,02 s |
| **FastPitch** | WaveGlow | 4,08 | Tacotron 2: 3,95 | x | x | x | x |
| **FastPitch MultiSpeaker 57h** | WaveGlow | 4,07 | Tacotron 2: 3,71 | x | x | x | x |

Autoregressive Modelle wie Tacotron 2 und Transformer TTS (siehe Tabelle 3) erzielen zwar exzellente Qualität, haben allerdings den Nachteil, dass sie lange zum Generieren der Sprachausgabe brauchen, da sie schlecht parallelisierbar sind. Wenige Minuten Audio nehmen schnell Stunden für die Generierung in Anspruch [22]. Daher hat sich die Forschung 2019 und 2020 überwiegend darauf konzentriert Modelle zu finden, die deutlich schneller sind und ähnlich gute MOS Werte liefern, statt weiter an noch besserer Sprachqualität zu arbeiten. Weitere Nachteile dieser Modelle sind, dass sie zu Wortauslassungen oder auch Wiederholungen neigen, sowie sich beim Inferencing nicht-deterministisch verhalten.

**Parallel Tacotron** setzt direkt auf der Tacotron Architektur auf und versucht diese zu Parallelisieren [23]. Dadurch wird eine Beschleunigung um Faktor 10 erzielt, aber die MOS Einschätzung sinkt um 0,13. **Fastspeech** [22] schafft eine um Faktor 270 schnellere Generierung als Transformer TTS, wobei die subjektive Qualität nur um 0,04 MOS gegenüber Transformer TTS sinkt. Diese liegt aber in der Veröffentlichung nur bei 3,88 MOS gegenüber 4,41 MOS für die menschlichen Sprecher. Der Nachfolger [24] kann zu Transformer TTS aufschließen (MOS 3,79 gegenüber 4,27 Ground Truth) und liegt damit knapp vor Tacotron 2 mit 3,74 MOS. Neben der schnellen Geschwindigkeit der Sprachsynthese sinkt mit **FastSpeech 2** auch die Trainingszeit, um die entsprechende Qualität zu erzeugen von 38,64 h für Transformer TTS auf 16,46 h. Das geht auch mit einer erheblichen Energieersparnis einher. **AlignTTS** [25] gelingt es schließlich sogar Transformer TTS mit einem MOS Wert von 4,05 um 0,03 MOS zu überbieten, während es bei der Zeit zur Generierung der Daten nur wenig langsamer als Fastspeech ist (0,18 s im Gegensatz zu 0,16 s zum Generieren einer Sekunde Audio).

Die anderen aktuellen Sprachsynthesemodelle vergleichen ihre Ergebnisse nicht mit Transformer TTS, sondern mit Tacotron 2, so dass unklar bleibt, ob sie nochmals besser als AlignTTS abschneiden. In Tabelle 2 findet sich eine Übersicht über die verschiedenen Modelle. GT steht für die Ground Truth, also den Vergleich mit dem menschlichen Sprecher. GT-MOS ist die Differenz zwischen der Ground Truth und dem MOS-Wert des Modells. Mel GT bezieht sich auf den MOS Wert, der mit dem Vocoder alleine auf Basis der aus der Ground Truth erzeugten Mel-Spektrogramme erreicht wurde. Leider wird dieser Wert nicht überall angegeben, genau wie die Geschwindigkeit der Spracherzeugung (Zeit pro 1 s erzeugtes Audio).

Analysiert man die Werte in Tabelle 2, so fällt auf, dass kein Modell so nahe an die Ground Truth herankommt wie Tacotron 2 und Transformer TTS, die beide WaveNet als Vocoder benutzen. Das ist insofern überraschend, als die meisten anderen WaveGlow benutzen und im WaveGlow Artikel dessen MOS um 0,08 besser ist als WaveNet. Vermutlich nutzen die Autoren von Tacotron 2 und Transformer TTS eine WaveNet Version, die mit deutlich mehr Trainingsdaten trainiert wurde, da in [8] ebenfalls eine Steigerung von 0,20 MOS berichtet wird, wenn die Trainingsdaten für WaveNet von 16 kHz, 8 bit und 25h auf 24 kHz, 16 bit und 65h verbessert werden.

Viele der effizienten Modelle benötigen ein aufwändiges Training, bei dem zuerst ein sog. *Lehrer* trainiert wird (z.B. TransformerTTS oder Tacotron 2) und dann davon ein *Student* abgeleitet wird, der deutlich schneller ist (siehe Tabelle 2). Dieser Prozess ist als Distillation bekannt und in seltenen Fällen übertreffen die *Studenten* nach dem Training den *Lehrer* sogar in der Qualität. **Flow-TTS** [26] ist eines der wenigen Modelle, das ohne diese *Distillation* auskommt und trotzdem non-autoregressive arbeitet und damit sehr schnell ist. Auch von der Qualität her kann es überzeugen, da es um 0,09 MOS besser als Tacotron 2 abschneidet und nach Tacotron 2 und Transformer TTS auch am nächsten an der Ground Truth liegt. **Glow-TTS** [27] fällt in die gleiche Kategorie, schneidet aber bzgl. des MOS-Wertes schlechter ab, obwohl es ebenfalls LJSpeech als Grundlage verwendet und Tacotron 2 deutlicher übertrifft als Flow-TTS.

Tabelle 3: Kategorisierung von Sprachsynthese Modellen nach Architektur [26]

| Autoregressive | Non-autoregressive | |
| --- | --- | --- |
| | Need distillation | No distillation |
| Tacotron [5] | ParaNet [15] | **Flow-TTS** |
| Tacotron 2 [6] | FastSpeech [14] | |
| Deep Voice 3 [7] | | |
| TransformerTTS [16] | | |
| WaveNet [9] | ParallelWaveNet [17] | WaveGlow [10] |
| WaveRNN [18] | ClariNet [8] | FloWaveNet [19] |
| LPCNet [20] | | |

**TalkNet** geht das Problem mit ausgelassenen Wörtern und Kontrolle der Geschwindigkeit der generierten Sprache an [28]. Es arbeitet mit CNNs, was für Sprachsynthese eher ungewöhnlich ist und erzielt damit gute Ergebnisse, die jedoch nicht ganz an Tacotron 2 heranreichen. Dafür liegt es in der Generierungsgeschwindigkeit selbst gegenüber Fastspeech noch einmal deutlich zu und kann in 16h auf einer V100 GPU trainiert werden. Für die proklamierten Verbesserungen bzgl. ausgelassener Wörter und Sprechgeschwindigkeit werden leider keine Belege genannt.

**FastPitch** [29] ist insofern interessant, als es mit nur 57 h Multi-Speaker Daten (3 Sprecherinnen) die gleiche Qualität erzeugt wie mit den 24 h Single-speaker Daten aus dem LJSpeech Datensatz. Der MOS Wert von 4,07 liegt deutlich über den berichteten MOS Werten für Multi-Speaker Training aus anderen Veröffentlichungen. Auch im Single-speaker Vergleich schneidet es besser ab, als das zugrundeliegende Fastspeech 2. Der Trainingsaufwand wird für LJSpeech mit 16 GPU-Stunden bis zur Konvergenz und 44 Stunden bis zum vollständigen Training angegeben, was ebenfalls ein guter Wert ist. Interessanterweise arbeitet Nvidia, von denen FastPitch stammt, parallel auch noch an anderen Modellen zur Sprachsynthese, wie z.B. **Flowtron** [30]. Dieses basiert nicht auf Fastspeech 2 und legt mehr Wert auf die Kontrolle linguistischer Parameter wie Tonhöhe, Sprechgeschwindigkeit, Kadenz und Betonung. Das Training erfolgte mit demselben Multi-Speaker Datensatz wie FastPitch, erzielte bei Flowtron aber nur mittelmäßige MOS Werte von 3,66.

Einen interessanten Zwitter stellt **SpeedySpeech** dar [31]. Es benutzt auch einen Teacher-Student Ansatz, soll jedoch schneller zu trainieren sein, weil der Teacher, welcher von DeepVoice 3 abgeleitet ist, für den Student nur zum Alignment zwischen Text und Mel-Spektrogramm dient. Daher soll das gesamte Modell mit LJSpeech in 40 GPU Stunden auf einer Nvidia GTX 1080 GPU mit 8 GB RAM trainierbar sein, was ein guter Wert wäre. Die Qualität der Ergebnisse wurde leider nur in Form des weniger häufig verwendeten MUSHRA Index ermittelt. Auf dieser Skala (bis 100) schnitt SpeedySpeech mit 75,24 deutlich besser ab als Tacotron 2 mit 62,82. In beiden Fällen wurde MelGAN als Vocoder benutzt.

## Datensätze

Für das TTS Training muss zunächst zwischen **Single- und Multi-Speaker Datensätzen** unterschieden werden. Die besten Ergebnisse werden bisher mit Single-Speaker Daten erzielt, bei denen nur ein(e) SprecherIn das komplette Audiomaterial eingesprochen hat, am besten mit dem immer gleichen Mikrofon und unter konstanten Umgebungsbedingungen ohne Nebengeräusche. Will man dagegen mehrere unterschiedliche Stimmen mit dem gleichen Modell realisieren, so bieten sich Multi-Speaker Datensätze an. Dort wird neben der Audio-Datei auch noch eine Sprecher-ID für das Training angegeben. Man benötigt zudem deutlich mehr Trainingsdaten. Beim Inferencing kann dann die Speaker-ID angegeben und so gezielt eine Stimme reproduziert werden.

Am häufigsten wird der **LJSpeech** Datensatz eingesetzt, für den eine einzelne professionelle Sprecherin (Linda Johnson) 24 Stunden Audiomaterial in Studioqualität aufgenommen hat[2]. Die Daten stammen aus dem LibriVox Projekt, das frei verfügbare Hörbücher sammelt. Die einzelnen Audio-Dateien haben eine Länge von 1,1-10,1 Sekunden (Mittelwert: 6,6) und beinhalten 13.821 unterschiedliche Wörter, bei 225.715 Wörtern insgesamt und durchschnittlich 17 Wörtern pro Datei.

Neben LJSpeech wird im Englischen oft **LibriTTS**[3] eingesetzt, ein Multi-Speaker Datensatz mit 585 Stunden englischer Sprache von 2.456 Sprechern bei 24 kHz Samplingrate, das ursprünglich ebenfalls aus dem LibriVox Projekt stammen. Zielführender als sehr viele Sprecher mit wenigen Stunden Audio pro Sprecher scheint es aber zu sein, mit nur ein paar Dutzend Sprecher zu arbeiten, die jeweils 5-10 Stunden Audiomaterial beitragen. FastPitch arbeitet bei seinem 57h Datensatz sogar mit nur 3 Sprecherinnen [29].

**LibriVox** enthält neben den fast 29.000 englischen Bücher auch 2.400 deutsche. Daraus speist sich auch der Datensatz des **Münchner AI-Labs**[4], die 69 h der Sprecherin „Ramona Deininger-Schnabel", 40 h des Sprechers Karlsson und 29 h bzw. 24 h der Sprecherinnen „Eva K." und „Rebecca Braunert-Plunkett" für die Sprachsynthese aufbereitet haben. Diese liegen jedoch nur mit 16 kHz Sampling Rate vor (außer Rebecca mit 32 kHz), was zu deutlichen Qualitätseinbußen im Vergleich zu 22 oder 24 kHz führt. Die 16 h Audiodaten für Deutsch aus dem Multilingualen **CSS10**[5] Datensatz stammen ebenfalls aus dem LibriVox Projekt und verwenden drei Bücher der Sprecherin Hokuspokus.

Auch der **Spoken Wikipedia Corpus**[6] könnte als Grundlage für einen TTS Datensatz geeignet sein. Vermutlich wird er aber nur für ein Multi-speaker Training sinnvoll zu nutzen sein, da dort 249 h deutsche Sprachdaten von insgesamt 339 Sprechern vorhanden sind. Die Daten von **Mozilla Common Voice**[7] sind überwiegend qualitativ so schlecht und von vielen Sprechern mit wenig Audiomaterial statt wenigen mit viel Audiomaterial, dass sie nicht ohne aufwändiges Aussortieren für die Sprachsynthese verwendet werden können und selbst im Vergleich zu anderen Datensätzen für Spracherkennung schlecht abschneiden [32].

Nach Vorbild von LJSpeech gibt es einen bisher wissenschaftlich noch gar nicht dokumentierten Datensatz von Thorsten Müller[8] namens **OGVD** (Open German Voice Dataset). Dieser hat selbst 22 h Audiomaterial eingesprochen und dabei extrem deutlich, beinahe schon überbetont gesprochen. Daher konvergiert das Training mit diesem Datensatz schneller als mit anderen Datensätzen. Das Resultat klingt allerdings vermutlich aus denselben

---

[2] https://keithito.com/LJ-Speech-Dataset/
[3] https://openslr.org/60/
[4] https://www.caito.de/2019/01/the-m-ailabs-speech-dataset/
[5] https://github.com/Kyubyong/CSS10
[6] https://nats.gitlab.io/swc/
[7] https://commonvoice.mozilla.org/de/datasets
[8] https://github.com/thorstenMueller/deep-learning-german-tts/

Gründen zwar sehr deutlich, aber vom Redefluss her etwas merkwürdig. Vortrainierte Modelle mit dem OGVD gibt es bei Coqui und Silero[9].

Obwohl es viele Sprachressourcen im Deutschen gibt, wie z.B. im **Bavarian Archive for Speech Signals**[10] oder in der „European Research Infrastructure for Language Resources and Technology" (**CLARIN**)[11], sind wenig brauchbare Daten für Sprachsynthese mit Deep Learning zu finden.

## Zusammenfassung und Diskussion

Insgesamt gibt es eine Reihe gut funktionierender Modelle und es scheint eher eine Frage des Finetunings der Hyperparameter, sowie der Menge und Qualität der Trainingsdaten zu sein, die darüber entscheidet wie weit die Modelle es schaffen über die Grenze von 4,0 MOS zu klettern. Die größeren Unterschiede ergeben sich darin, wie einfach und schnell das Netz trainiert werden kann. Das lässt sich leider nur selten aus den Veröffentlichungen ablesen. Die Geschwindigkeit der Generierung ist je nach Anwendungsfall entscheidend oder nebensächlich. Hier unterscheiden sich die Modelle deutlich. Für gute Sprachsynthese auf Deutsch muss dem Aufbau eines eigenen Datensatz an hochwertigem Audiomaterial zum Training eine hohe Priorität eingeräumt werden, da er einer der ausschlaggebendsten Punkte für die Qualität der generierten Sprache ist. Ein Beleg dafür findet sich in Tabelle 2, wenn man die Tacotron 2 MOS Werte in der Zeile mit Flowtron und der mit Tacotron 2 vergleicht. Im Originalpaper erreicht Tacotron 2 hervorragende 4,53 MOS, bei einer ebenfalls sehr guten Einschätzung der Ground Truth von 4,58, während in der Flowtron Veröffentlichung Tacotron 2 nur noch 3,52 MOS erreicht, was wohl nicht nur, aber sicherlich zu einem erheblichen Teil an der geringeren Qualität der Ground Truth liegt, die hier nur auf 4,27 eingeschätzt wird. Die Open Source verfügbare Sprachsynthese Software basiert noch weitgehend auf älteren Ansätzen ohne deep learning. In Deutschland ist v.a. Mary-TTS vom DFKI zu nennen [34, 35], das eine ordentliche Qualität erzielt, ohne jedoch an die State-of-the-Art Ergebnisse wie Tacotron 2 mit WaveNet heranzureichen.

## Literatur

---

[9] https://github.com/snakers4/silero-models
[10] https://www.bas.uni-muenchen.de/Bas/
[11] https://www.clarin.eu/resource-families/spoken-corpora